\documentclass[journal,transmag]{IEEEtran}
\ifCLASSINFOpdf
\else
\fi
 \usepackage{wrapfig}

\usepackage{graphicx}
\usepackage[utf8]{inputenc} 
\usepackage[T1]{fontenc}    
\usepackage{hyperref}       
\usepackage{url}            
\usepackage{booktabs}       
\usepackage{amsfonts}       
\usepackage{nicefrac}       
\usepackage{microtype}      
\usepackage{psfrag}
\usepackage{amssymb}
\usepackage{amsbsy}
\usepackage{graphicx}
\usepackage{array}
\usepackage{multirow}
\usepackage{bm}
\usepackage{verbatim}
\usepackage{relsize}
\usepackage{algorithm}
\usepackage{algpseudocode}
\usepackage{tabularx}
 \usepackage{verbatim}
\usepackage{bm}
\usepackage{lipsum}
\usepackage{tabularx}
\usepackage{mwe} 
\usepackage[utf8]{inputenc} 
\usepackage[T1]{fontenc}    
\usepackage{hyperref}       
\usepackage{url}            
\usepackage{booktabs}       
\usepackage{nicefrac}       
\usepackage{subfigure}
\usepackage{microtype}      
\usepackage{times}
\usepackage{epsfig}
\usepackage{graphicx}
\usepackage{amsmath}
\usepackage{mathrsfs} 
\usepackage{amssymb}
   \usepackage[T1]{fontenc}
    \usepackage{mathpazo}

    \usepackage{graphicx}

\newcommand{\yy}{{\bf y}}

\newcommand{\ccx}{{\bf c}_x}
\newcommand{\ccy}{{\bf c}_y}
\newcommand{\cc}{{\bf c}}
\newcommand{\bbx}{{\bf b}_x}
\newcommand{\bby}{{\bf b}_y}

\usepackage{color}

\usepackage{caption}
\usepackage{wrapfig}
\usepackage{tabularx}
\usepackage{placeins}
\usepackage{dblfloatfix}

\begin{document}
%
\title{Hyperparameter Analysis for Derivative Compressive Sampling}


\author{\IEEEauthorblockN{Md Fazle Rabbi\\ \href{mailto:mrabbi@fordham.edu}{mrabbi@fordham.edu}}
\IEEEauthorblockA{ Fordham University, New York, NY, 10458 USA} 
}

\markboth{ISI REU Report, August 2021}%
{Shell \MakeLowercase{\textit{et al.}}: Bare Demo of IEEEtran.cls for Journals}
%



\IEEEtitleabstractindextext{%
\begin{abstract}
Derivative compressive sampling (DCS) is a signal reconstruction method  from  measurements of spatial gradient with sub-Nyquist sampling rate.  Applications of DCS include optical image reconstruction, photometric stereo, and shape-from-shading. In this work, we study sensitivity of DCS with respect to algorithmic hyperparameters using brute-force search algorithm. We perform experiments on a dataset of surface images and deduce guidelines for the user to setup values for the hyperparameters for improved signal recovery performance.
\end{abstract}

}

\maketitle

\IEEEdisplaynontitleabstractindextext

%
\IEEEpeerreviewmaketitle

\section{Introduction}
\IEEEPARstart{T}{he} standard setting to measure continuous signals is to sample the source signal with rate more than the Nyquist rate which allows to reconstruct band-limited signals uniquely~\cite{landau1967sampling}. However, sampling rate might be limited in practical setting.  Reconstruction of source signals from sub-Nyquist measurements has been explored  in various problem settings, including compressive sampling~\cite{4,13,qq4,rostami2013compressedfield,qiao2020deep} and blind source separation~\cite{diamantaras2008blind,kim2009underdetermined,rostami2011blind,lin2020nonnegative}. Compressive sampling is a setting to reconstruct a source signal by solving underdetermined linear systems when each measurement is a linear combination of singal samples, while blind source separation considers that a mixture of several source signals is measured and the goal is to separate the source signal samples. In this work, we perform sensitivity analysis for a special case of compressive sampling, called Derivative compressive sensing (DCS)~\cite{rostami2012image}. DCS is a method for reconstruction of a source signal from its gradient field measurements. Its applications include surface reconstruction \cite{rostami2015surface,rostami2012gradient}, diffusion field reconstruction~\cite{rostami2013compressed}, and optical image reconstruction \cite{rostami2012image}. In these applications, the goal is to recover a two dimensional signal $(x,y)$  from the measurements of
its spatial gradient. First, the gradient field $(z_x,z_y)$ is
measured at each point $(x,y)$ and then the original signal is obtained through integration of the gradient field through
the solution of a Poisson equation.

When the hardware sampling rate is limited, DSC can be used to recover the original signal from gradient measurements more efficiently than the classic CS approach because the gradient signal properties are used a side information for improvement. The rest of this report is organized as follows. Section II briefly summarizes a background on compressed sensing. In Sections III, we explain the DCS reconstruction algorithm. In section IV, we describe our hyperparamter study approach and present experimental results  to study sensitivity aspects of DCS. Finally,
Section V concludes the report.

\section{Compressive Sampling}
CS takes advantage of sparsity as a prior on source signals to recover them as a unique solution to an undetermined linear system. Sparsity as a prior has been found to be practical for a wide range of applications~\cite{re4,purisha2017controlled,qiu2019jointly,kolouri2018joint,rostami2020using,ma2020dynamic}. 
Consider a source signal $\textbf{x} \in \mathbb{R}^n$ that can be represented sparsely with respect to a discrete basis $W\in\mathbb{R}^{n\times n}$, i.e., $\textbf{x}=W\textbf{c}$, where $\textbf{c}$ is a sparse signal, i.e., many of its elements are equal to zero. The theory of CS states that a  sparse signal  can be recovered from $m<n$ linear measurements:
\begin{equation} \label{2}
\yy = \Psi \textbf{x}+\textbf{n},
\end{equation}
where $\yy \in \mathbb{R}^m$. The matrix $\Psi \in \mathbb{R}^{m\times n}$ is called the {\em sensing matrix} and $\textbf{n}$ denotes measurement noise, e.g., Gaussian noise.
If $\textbf{c}$, then the source signal can be recovered as a unique solution to Eq.~\eqref{2} by solving the following optimization problem:
\begin{equation}\label{3}
\textbf{c}^* = \arg \min_{\textbf{c}^\prime}  \| \textbf{c}^\prime \|_1+\lambda \| \Phi \textbf{c}^\prime - \yy \|_2^2,
\end{equation}
where $\lambda > 0$ is a regularization hyperparameter. In order to recover the signal with small error, the regularization hyperparameter needs to be tuned properly. Several algorithms to  solve \eqref{3} \cite{33,cc10,cc11}. After recovering the source signal through solving Eq.~\eqref{3}, we can recover the source signal as: $\textbf{x}^*=W\textbf{c}^*$.

 \section{Derivative Compressive Sampling}
In some applications of CS, we have a secondary prior on the source signal in addition to saprsity~\cite{rostami2013compressed}. Let $z(x,y)$  to be a two-dimensional source signal that is measured through  the samples of its partial derivatives ${\bf z}_x\in\mathbb{R}^{n}$ and ${\bf z}_y\in\mathbb{R}^{n}$. The derivative samples are concatenated to form two column vectors. The measurement vectors  ${\bf b}_x\in\mathbb{R}^{m}$ and ${\bf b}_y\in\mathbb{R}^{m}$ corresponding to  ${\bf z}_x$ and ${\bf z}_y$, respectively. Following the CS setting, we assume these two vector are obtained according to ${\bf b}_x = \Psi_x \, {\bf z}_x$ and ${\bf b}_y = \Psi_y \, {\bf z}_y$, where $\Psi_x\in\mathbb{R}^{m\times n}$ and $\Psi_y\in\mathbb{R}^{m\times n}$. We also assumed there are two sparse  representations ${\bf c}_x$ and ${\bf c}_y$ such that ${\bf z}_x = W \, {\bf c}_x$ and ${\bf z}_y = W \, {\bf c}_y$.  Hence,  we can solve for ${\bf c}_x$ and ${\bf c}_y$   according to:
\begin{equation}\label{ekk1}
\ccx^\ast = \arg \min_{\ccx^\prime} \left\{ \frac{1}{2} \| \Psi_x W \ccx^\prime - \bbx \|_2^2  + \lambda \| \ccx^\prime \|_1 \right\}
\end{equation}
and
\begin{equation}\label{ekk2}
\ccy^\ast = \arg \min_{\ccy^\prime} \left\{ \frac{1}{2} \| \Psi_y W \ccy^\prime - \bby \|_2^2  + \lambda \| \ccy^\prime \|_1 \right\}.
\end{equation}
If we set $\cc = [\ccx^T, \ccy^T]^T$, $\yy = [\bbx^T, \bby^T]^T$, and $\Phi = \mbox{diag} \{\Psi_x W, \Psi_y W\}$, we can combine Eq.~\eqref{ekk1} and Eq.~\eqref{ekk2} into a one problem as:
\begin{equation} \label{E4}
\cc^\ast = \arg \min_{\cc^\prime} \left\{ \frac{1}{2} \| \Phi \cc^\prime - \yy \|_2^2  + \lambda \| \cc^\prime \|_1 \right\}.
\end{equation}

The DCS is to benefit from side information which arises from the fact that for the source signal $z(x,y)$, we have the following relation:
\begin{equation} \label{E4qw}
\frac{\partial^2 z}{\partial x \, \partial y}=\frac{\partial^2 z}{\partial y \, \partial x}.
\end{equation}
Now if we let $D_x$ and $D_y$ denote the matrices of discrete partial differences, we can conclude for Eq.~\eqref{E4qw}:
\begin{equation} \label{25}
D_x \textbf{z}_y = D_y \textbf{z}_x.
\end{equation}

Now if we let $B := D_y W T_x - D_x W T_y$, with $T_x$ and $T_y$ defined as ${\bf c}_x = T_x \, {\bf c}$ and ${\bf c}_y = T_y \, {\bf c}$. Then, we have  $B {\bf c} = 0$ and deduce
\begin{equation} \label{26}
\yy' = \Phi' \textbf{c}+\textbf{n}',
\end{equation}
where $\Phi' : = [\Phi^T, B^T]^T$, ${\bf y}' = [{\bf y}^T, 0^T]^T$, and ${\bf n}' = [{\bf n}^T, 0^T]^T$.

To solve Eq.~\eqref{26}, we can formulate the following constrained optimization problem:
\begin{align} \label{E5}
&\cc^\ast = \arg \min_{\cc^\prime} \left\{ \frac{1}{2} \| \Phi  \cc^\prime - \yy \|_2^2  + \lambda \| \cc^\prime \|_1 \right\}, \\
&\mbox{s.t.  } B \cc^\prime = 0. \notag
\end{align}
This can be solved by the augmented Lagrangian methods~\cite{cc13,hao2016testing} in an iterative scheme:
\begin{align} \label{E6}
\cc^{(t+1)} = \arg \min_{{\bf c}'} &\Big\{ \frac{1}{2} \| \Phi \cc^\prime - \yy \|_2^2  + \notag \\
+ &\lambda \| \cc^\prime \|_1 + \frac{\delta}{2} \| B \cc^\prime + p^{(t)} \|_2^2 \Big\} \\
p^{(t+1)} = p^{(t)} + B &\cc^{(t+1)}, \notag
\end{align}
where $p^{(t)}$ is a vector of   Lagrange multipliers, $\delta > 0$ is a penalty hyperparamter, and $t$ stands for the iteration index.  The performance of the DCS recovery also depends on the value for the hyperparamter $\delta$. Finally, given the optimal solution ${\bf c}^\ast$, the   measurements of partial derivatives can be recovered as ${\bf z}_x = W \, T_x \, {\bf c}$ and ${\bf z}_y = W \, T_y \, {\bf c}$, which can be recover the original signal $z(\cdot,\cdot)$ via solving the Poisson equation~\cite{rostami2012image}.

  \section{Hyperparamter Analysis}
 
 The performance of the DCS recovery algorithm depends on tuning the hyperparameters $\delta$ and $\lambda$. Tuning the values for these hyperparameters is important for the best recovery quality but the challenge is that in practice only the noisy measurements are accessible.  Our goal is to use brute-search algorithm to deduce guidelines to set up values for the hyperparameters~\cite{grothendieck2013non}. To this end, we focus on the problem surface reconstruction using DCS~\cite{rostami2015surface,rostami2012gradient}. We select a dataset of surface images. For a given surface image ${\bf z}$, we then synthetically generate noisy versions of its gradient measurements using a sampling matrix. We can apply DCS algorithm to recover the source signal using the noisy gradient measurements for fixed values of the hyperparameters $\lambda$ and $\delta$.
 Since the original surface image is available as an ideal reference image, we can compute the quality of reconstruction by comparing the DCS algorithm recovery and the original reference images. We can implement brute-search by varying  the hyperparameters $\lambda$ and $\delta$ values and compute the reconstruction quality for all instance. We can then find the optimal values for the hyperparameters $\lambda$ and $\delta$ that leads to the best performance. By doing so on several images, we can come up with guidelines for users to set up the values for the hyperparameters
 
 \subsection{Experiments}
 
  In our experiments, we have used brute-search algorithm to study the effect of values for the hyperparameters on the surface recovery problem.
 
  \subsubsection{Experimental Setup}
\begin{figure}
\centering
\subfigure[]{
\includegraphics[height= 2.7cm,width = 2.7cm]{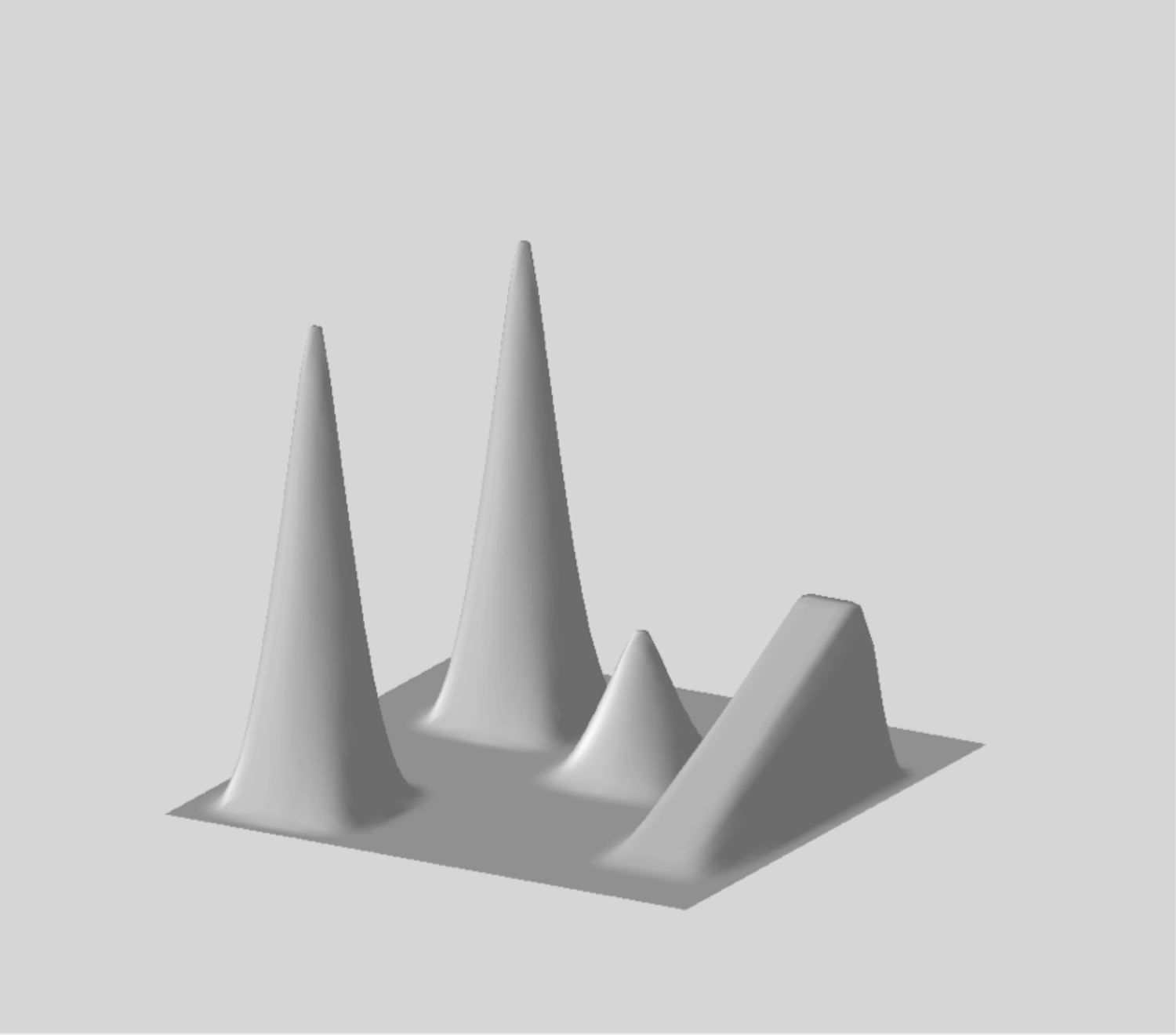}
\label{fig3:tabsubfig1}
}
\hspace{-4mm}
\subfigure[]{
\includegraphics[height= 2.7cm,width = 2.7cm]{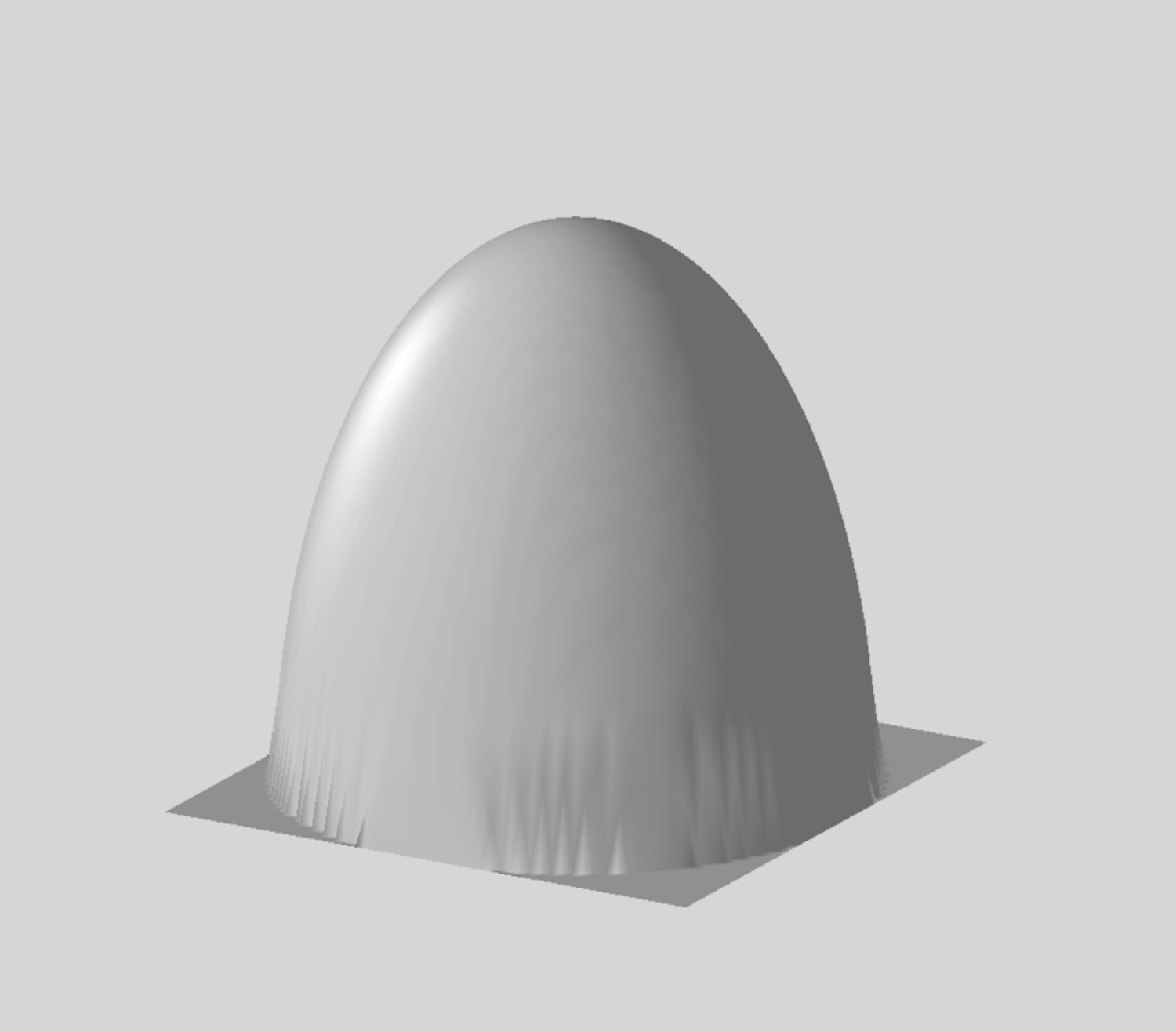}
\label{fig3:tabsubfig2}
}

\hspace{-4mm}
\subfigure[]{
\includegraphics[height= 2.7cm,width =2.7cm]{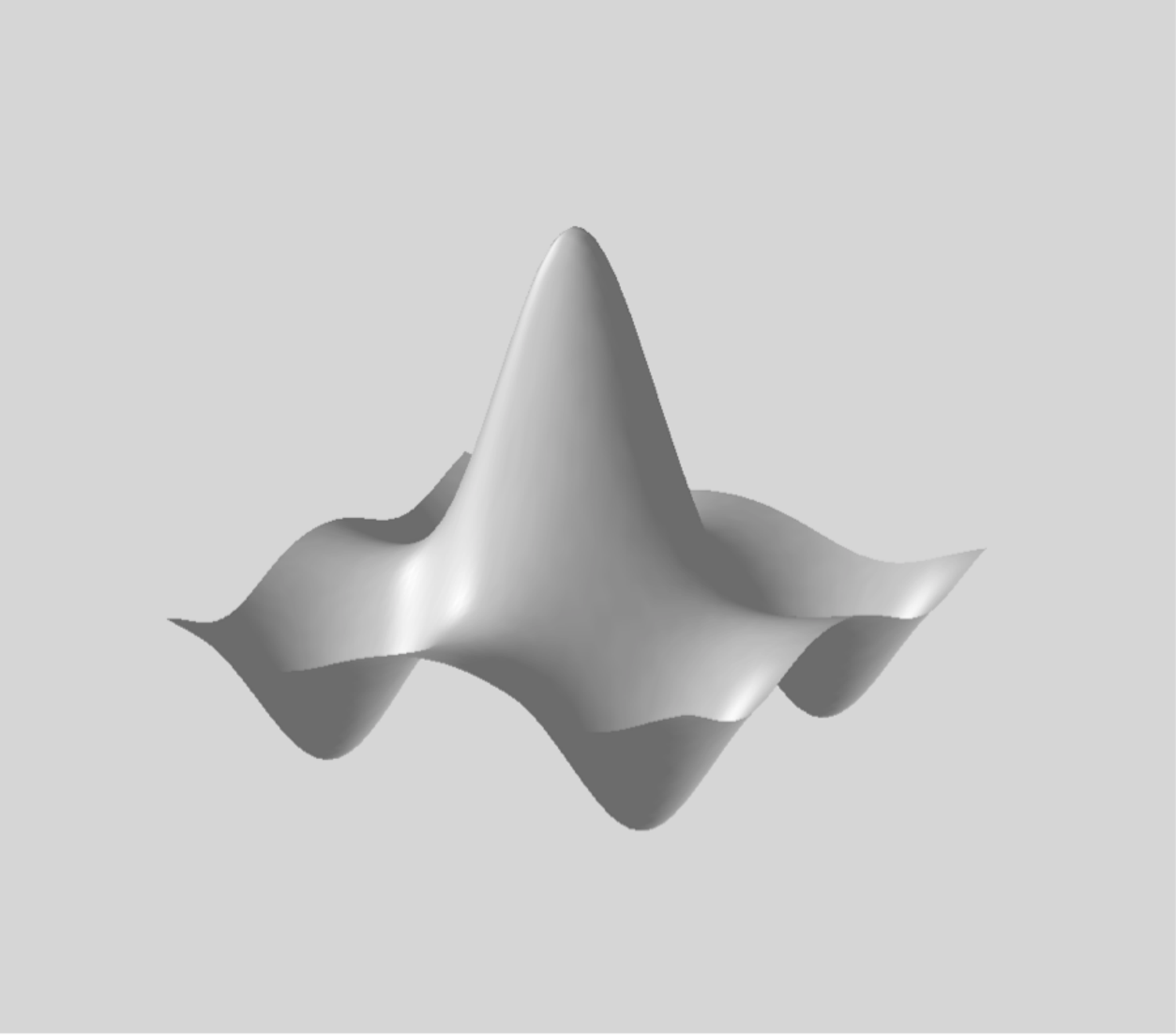}
\label{fig3:tabsubfig3}
}  
\\
\subfigure[]{
\includegraphics[height= 3.9844cm,width = 8.6cm]{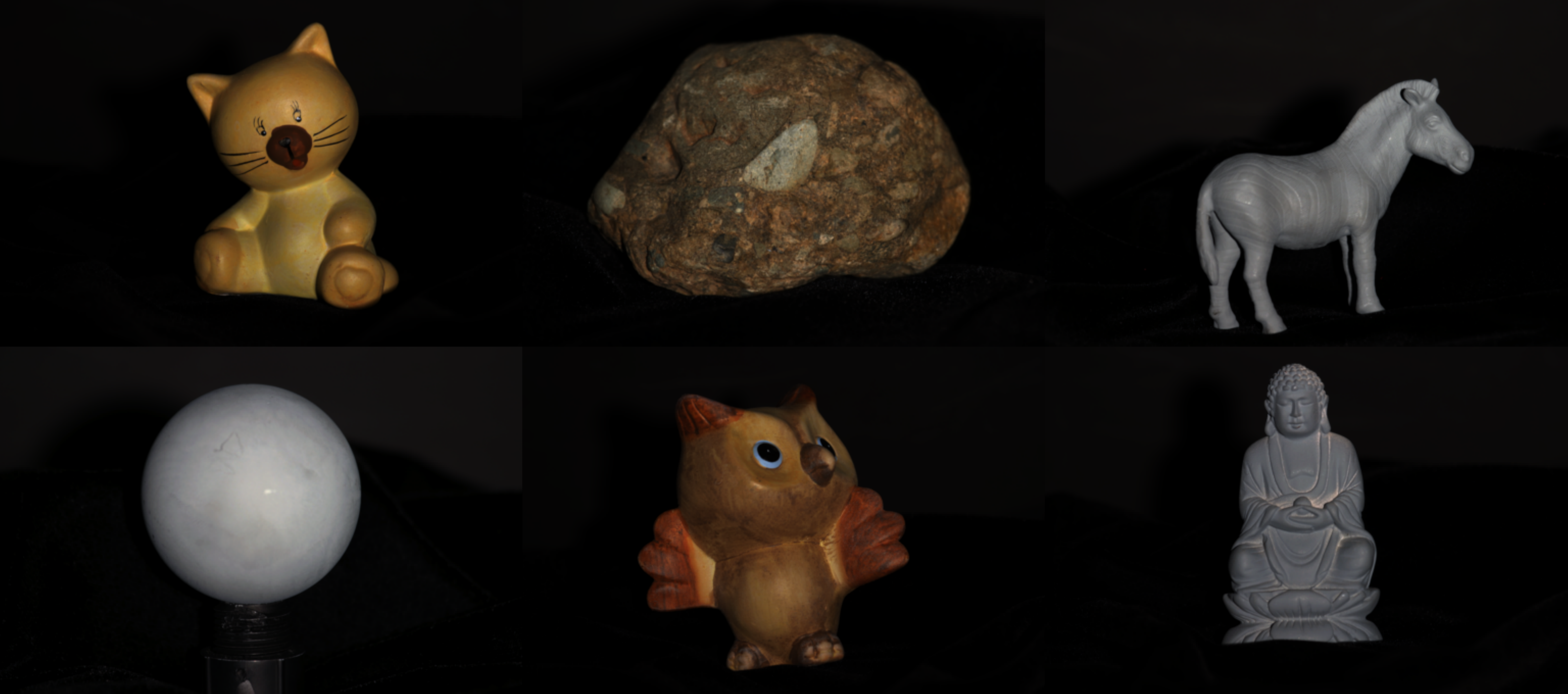}
\label{fig3:tabsubfig4}
}
\caption{"Nine synthetic surfaces (a) Ramp-peak (b) Sphere, (c) Peak-valley, and (d) sample photos of 6 real objects. (Top from left to right: Cat, Rock, and Horse. Bottom from left to right: Gray, Owl, and Buddha)"}
\label{fig3}
\end{figure}

\textbf{Surfaces}:  "We have used nine surfaces in our experiments. Figure~\ref{fig3} shows these nine surface. As it can be seen, these surfaces include both synthetic and also semi-real world objects. They also have different levels of details which make them diverse and suitable to draw conclusions from our exploration."
 
 \begin{table*}[t]
 \centering 
\begin{tabular}{l|lllllllll}
Surface   & Ramp-peak & Sphere & Peak-valley &  Cat& Rock & Horse & Gray & Owl & Buddha \\
\hline
\hline
$\lambda$ & 0.0081 & 0.00006 & 0.00082 & 0.03312 & 0.01242 & 0.0622 & 0.00126 & 0.0612 & 0.04312  \\
\hline
$\delta$  & 2.74 & 2.80 & 4.62 & 4.11 & 3.80 & 4.71 & 3.23 & 2.41 & 3.54 \\
\hline
\end{tabular}
\caption{Optimal  values for the hyperparameters for the Gaussian noise.}
\label{tab:table1}
\end{table*}

\begin{table*}[t]
 \centering 
\begin{tabular}{l|lllllllll}
Surface   & Ramp-peak & Sphere & Peak-valley &  Cat& Rock & Horse & Gray & Owl & Buddha \\
\hline
\hline
$\lambda$ & 0.0062 & 0.00007 & 0.00042 & 0.0432 & 0.03213 & 0.0712 & 0.00343 & 0.07101 & 0.0523 \\
\hline
$\delta$ & 1.95 & 1.53 & 4.00 & 2.02 & 3.61 & 1.55 & 1.15 & 3.93 & 4.43 \\
\hline
\end{tabular}
\caption{Optimal  values for the hyperparameters for the Laplace noise.}
\label{tab:table2}
\end{table*}

\begin{table*}[t]
 \centering 
\begin{tabular}{l|lllllllll}
Surface   & Ramp-peak & Sphere & Peak-valley &  Cat& Rock & Horse & Gray & Owl & Buddha \\
\hline
\hline
$\lambda$ & 0 & 0 & 0 & 0 & 0 & 0 & 0 & 0 & 0 \\
\hline
$\delta$  & 3.20 & 3.74 & 4.20 & 0.10 & 0.59 & 2.20 & 3.20 & 2.60 & 1.08 \\
\hline
\end{tabular}
\caption{Optimal  values for the hyperparameters for the salt and pepper noise.}
\label{tab:table3}
\end{table*}

\textbf{Noise types:} we have considered three noise types in our experiments: Gaussian noise, Laplace noise, and salt and pepper noise. These noise emerge in practice conditioned on the type of measurement device we use. 
 Gaussian noise usually is caused from
 sensors by poor illumination, high temperature, and transmission. Laplace noise
 is an additive noise which emerges in cryptography applications. The salt and pepper noise can be caused  by sharp disturbances in the source signal. 
We explore these three noise types to broaden our exploration.

\textbf{Search values:} in our experiments, we have considered the hyperparameter values in the brute-search algorithm to be $\delta= [0.1, 1 , 2 , 5, 10]$ and $\lambda = [10^{-5}, 10^{-4}, 10^{-3}, 10^{-2}, 10^{-1}, 1, 10]$. We have also used the signal-to-noise (SNR) ratio to measure the quality of the reconstructed surface against the reference surface. For fixed values for the hyperparameters, we apply the DCS algorithm ten times and compute SNR ten times and report the average optimal values for $\delta$ and $\lambda$ to cancel out the random nature of the noise.

 \subsubsection{Results and Analysis}
  Tables~1--3 represent our results. We can conclude from these results that as long as the hyperparameter $\lambda$ value is close to 0 or 0, DCS recovery leads to the highest SNR regardless of the noise type. As for the hyperparameter $\delta$ value, we couldn't find any single value that will work for all nine figures and all the three primary noises. But, we were able to deduce to this much to set the ranges for hyperparameter $\delta$ value for three primary noises. The hyperparameter $\delta$ range for Gaussian noise from 2.74 to 4.71, Laplace noise from 1.15 to 4.43, and Salt and Pepper noise from 0.10 to 4.20 lead to relatively good performance. Even though we found the optimal ranges for the hyperparameter $\delta$ from our experiment, but we couldn't deduce that a single value for $\delta$ leads to the best performance for all the scenarios.  We note however that the DCS recovery sensitivity to the value of $\delta$ quite small in the ranges that we found leads good recovery performance.

  \section{Conclusions}
In this project, we studied the sensitivity of DCS with respect to optimization hyperparameter to deduce guidelines for the user to tune the hyperparameters. We used the brute-force search technique to tune the hyperparameters on a collection of surface images for three primary noise types. Our experiments help the user to set up relatively suitable values for the two important hyperparameters after applying the brute-force search technique to tune the hyperparameters under the three primary noises Gaussian, Laplace, and Salt and Pepper. We were able to conclude optimal hyperparameter for $\lambda$ is 0 based on our experiment. But, unfortunately, we weren't able to conclude any such conclusion for hyperparameter $\delta$. The only thing we were able to conclude for hyperparameter $\delta$ is its ranges for each of the three primary noises. So, for Gaussian noise, the hyperparameter $\delta$ value ranges from 2.74 to 4.71, for Laplace noise, the hyperparameter $\delta$ value ranges from 1.15 to 4.43, and for hyperparameter $\delta$ value ranges from 0.10 to 4.20. Finally, from our research experiment, we were able to set the user guidelines for both hyperparameters $\lambda$ and $\delta$ that can be easily used to highest reconstructed signal for any images.

\end{document}